\pdfoutput=1

\documentclass[11pt]{article}

\usepackage{acl}
\usepackage{graphicx}
\usepackage{times}
\usepackage{latexsym}
\usepackage{amsmath}
\usepackage[T1]{fontenc}
\usepackage[utf8]{inputenc}
\usepackage{url}

\usepackage{microtype}

\usepackage{booktabs}
\usepackage[autolanguage]{numprint}

\usepackage{adjustbox}
\usepackage{changepage}  
\usepackage{paralist}  
\usepackage{amssymb}
\usepackage{array}
\usepackage{multirow}

\title{Using Distributional Principles for the Semantic Study of Contextual Language Models}

\author{Olivier Ferret\\
  Université Paris-Saclay, CEA, List, F-91120, Palaiseau, France\\
  {\tt olivier.ferret@cea.fr}
}

\begin{document}
\maketitle

\begin{abstract}
Many studies were recently done for investigating the properties of contextual language models but surprisingly, only a few of them consider the properties of these models in terms of semantic similarity. In this article, we first focus on these properties for English by exploiting the distributional principle of substitution as a probing mechanism in the controlled context of SemCor and  WordNet paradigmatic relations. Then, we propose to adapt the same method to a more open setting for characterizing the differences between static and contextual language models.
\end{abstract}

\section{Introduction}

The introduction of contextual word embeddings such as ELMo \citep{peters-naacl-18} or BERT \citep{devlin-naacl-19} is considered as a major breakthrough in Natural Language Processing, especially for classification or sequence labeling tasks that can be addressed by supervised machine learning approaches. However, this kind of embeddings also represents a significant change from the viewpoint of distributional semantics. While previous approaches, including static word embeddings such as Skip-gram or CBOW models \citep{mikolov-iclr-13}, built the distributional representation of a word by cumulating the contexts in which it was found, the neural language models such as ELMo or BERT produce a representation for each occurrence of the word.

While this specificity is not a difficulty -- and even more so an advantage -- in the context of a supervised classification or sequence labeling task, it is a problem in contexts requiring a distributional representation of words at the type level and not only at the token level. This is the case for the evaluation of the semantic similarity of words, which is a classical tool for investigating the semantic properties of distributional representations through datasets, such as Simlex-99 \citep{hill-cl-15}, built for evaluating the correlation of the similarity computed from these representations with human judgments.

As illustrated by \citet{rogers-tacl-20}, the properties of contextual word embeddings have been the focus of many recent studies, especially in the case of BERT. Surprisingly, only a few of them consider the semantic properties of these models. Moreover, most of them are performed at the word type level, which involves building representations of words from the representation of their occurrences, which is neither a straightforward nor a neutral operation.

In this article, we propose to exploit the distributional principles to investigate the semantic properties of contextual word embeddings in terms of paradigmatic semantic relations without the requirement to build representations at the word type level. More precisely, we present the following main contributions:

\begin{itemize}
 \item we show that the distributional hypothesis can be applied at the word token level with contextual word embeddings and that ELMo and BERT exhibit specific semantic properties falling more into the scope of semantic similarity than semantic relatedness;
 \item we show how the transposition of this method to the word type level can be used to study the differences between contextual and static word embeddings.
\end{itemize}

\section{Related work}

\subsection{Distributional semantic similarity}

The work on distributional semantics is closely linked to the notion of semantic similarity since one major criterion for judging the quality of a distributional representation is its ability to account for semantic similarity, either by its correlation with direct human judgment \citep{rubenstein-cam-65} or more indirectly through the extraction of semantic similarity relations \citep{grefenstette-book-96,landauer-psy_review-97}. In our work, we focus more particularly on the second option, which is historically linked to the notion of distributional thesaurus: the semantic similarity relations of a word are extracted by retrieving its nearest neighbors according to the similarity of their distributional representations \citep{grefenstette-book-94,lin-acl_coling-98,curran-siglex-02}. This perspective was dominant for count-based approaches \citep{weeds-coling-04,ferret-lrec-10,padro-emnlp-14} but is also represented among studies on static word embeddings \citep{schnabel-emnlp-15,antoniak-tacl-18,ferret-acl-18}. In this article, we extend the use of this paradigm to the exploration of the semantic properties of contextual embeddings.

\subsection{Semantic Study of Contextual Word Embeddings}

In \citep{rogers-tacl-20}, the semantic studies of BERT-like models mainly point out work focusing on semantic roles in sentences, either in the context of semantic role labeling \citep{tenney-iclr-19} or tasks close to lexical substitution \cite{ettinger-tacl-20,gari_soler-starsem-19}. However, the work of \citet{vulic-emnlp-20} clearly presents the largest set of experiments concerning the semantic properties of BERT and RoBERTa models. It includes experiments concerning their correlation with human judgment in terms of semantic similarity or their ability to classify word pairs according to different types of relations \cite{chersoni-cogalex-16,xiang-cogalex-20}. Other studies also considered contextual embeddings from a semantic viewpoint but in more specific contexts: the impact of their training objectives \citep{mickus-scl-20}, their level of contextualization \citep{ethayarajh-emnlp-19}, their possible biases \citep{bommasani-acl-20}, their ability to represent word senses \citep{coenen-neurips-19}, to build representations for rare words \citep{schick-aaai-20}, to account for selectional preferences \citep{metheniti-coling-20} or to interpret logical metonymy \citep{rambelli-aacl-20}. Finally, \citep{chronis-conll-20} is linked to the representation of word senses through the notion of prototype but mainly applies it for  characterizing semantic similarity versus semantic relatedness and abstractness versus concreteness. While it has links with some of these studies, our work proposes a specific method, based on the distributional hypothesis, that aims at characterizing contextual word embeddings by focusing on paradigmatic relations observed at the word token level without the necessity to build representations at the word type level.

\section{Principles and Methodology}\label{sec:distr_hyp}

 With contextual embedding models, the similarity between embeddings can only be computed for words in context, i.e. occurrences of words. More precisely, the embedding built for an occurrence of a word with this kind of model integrates two dimensions: one dimension, resulting from the training on a language modeling task, accounts for the aggregation of all the contexts of the occurrences of the word, as for all distributional models; the second dimension takes into account the local context of the considered occurrence, i.e. its surrounding words.

In our case, we focus on the first dimension for evaluating the similarity between words, which requires being able to control the second one. We propose to perform this control in a very simple manner: for evaluating the similarity of two words, we put the two words in the same context, i.e. at the same position in the same sentence, and compute a word embedding for each of them by applying a contextual language model. In practice, this control is implemented by a substitution: for two words $w_1$ and $w_2$, a sentence $S_i$ containing an occurrence of $w_1$ is selected and a contextualized representation of $w_1$ is computed; then, $w_1$ is replaced by $w_2$ in sentence $S_i$ and a contextualized representation of $w_2$ is computed similarly as for $w_1$. By computing the similarity between the representations of $w_1$ and $w_2$, we can evaluate to what extent $w_1$ can be replaced by $w_2$ and, by following \citet{harris-word-54}'s principle of \emph{substitutability}, also obtain an evaluation of their semantic similarity. This evaluation is limited to the selected sentence but the application of this strategy to a set of sentences gives a more global picture of the semantic relationship between the two words. It should be noted that $w_1$ and $w_2$ do not have symmetric roles: we evaluate the similarity of $w_2$ with respect to $w1$ since the representation of $w_2$ is built in the context of a sentence in which $w_1$ initially occurs. In the rest of the paper, the terms \emph{key}, \emph{target}, and \emph{test sentence} will refer to $w_1$, $w_2$, and $S_i$ respectively.

We first exploit the presented principle of substitution for investigating the semantic properties of ELMo and BERT. 
More precisely, the idea is, for each word of a set of keys, to gather a set of targets such that each (key, target) pair is linked by a semantic relation of a known type. Given a test sentence containing an occurrence of the key, its targets can be ranked according to their similarity value with the key following the principle of substitution outlined previously. The resulting ranking indirectly ranks the types of semantic relations associated with the targets and gives some insights into the semantic properties of ELMo or BERT.

Our work focuses more particularly on paradigmatic relations, more precisely synonymy [\textsc{Syn}], hypernymy [\textsc{Hype}], hyponymy [\textsc{Hypo}], and cohyponymy [\textsc{Cohyp}] and adopts WordNet~3.0 as a reference resource for these types of relations. However, WordNet \citep{miller-ijl-90} is based on the notion of synset while sentences contain words and not word senses. To bypass this difficulty, we use as test sentences sentences from SemCor \citep{miller-hlt-93}, a subset of the Brown Corpus whose open class words were tagged with WordNet synsets. Hence, the sense of an occurrence of a key is known and the relation with the target depends on this sense, which makes the use of substitution particularly accurate. For instance, the second sense of the key \emph{disaster} (\emph{an event resulting in great loss and misfortune}) has the following test sentence in SemCor:

\begin{itemize}
\small
\item[{[}1{]}] Since the 1946 \textbf{disaster} there have been 15 tsunami in the Pacific, but only one was of any consequence.
\end{itemize}

\noindent This sense of \emph{disaster} has words such as \emph{cataclysm} or \emph{catastrophe} as synonyms, \emph{misfortune} as hypernym, \emph{tsunami} or \emph{meltdown} as hyponyms, and \emph{adversity} or \emph{misadventure} as cohyponyms. For evaluating to which extent ELMo for instance accounts more for synonymy than for hypernymy, the test sentence [1] is turned into sentences [2] and [3] by substituting a synonym or a hypernym for the key.

\begin{itemize}
\small
\item[{[}2{]}] Since the 1946 \textbf{catastrophe} there have been 15 tsunami in the Pacific, but only one was of any consequence.
\item[{[}3{]}] Since the 1946 \textbf{misfortune} there have been 15 tsunami in the Pacific, but only one was of any consequence.
\end{itemize}

\noindent The three sentences are encoded using ELMo and the representation of both the key in sentence [1] and the two targets in sentences [2] and [3] are extracted from ELMo's internal layers. Since ELMo has three layers -- one input non-contextual layer, layer~0, and two contextual layers, layer~1 and layer~2 -- we actually have three  representations for each word. In this example and for layer~1, the similarity, classically evaluated by the \emph{cosine} measure, between the representations of the key and the target synonym \emph{catastrophe} is equal to 0.89 while the similarity of the key and the target hypernym \emph{misfortune} is only equal to 0.55, giving a clear advantage to synonymy over hypernymy in that case. The process is the same for BERT, except that we have 12 layers (layer~1 to layer~12) in that case. A global picture of the semantic properties of ELMo or BERT is obtained by considering a significant number of keys and targets in the context of a large number of test sentences from SemCor. Moreover, we enrich this picture by considering the relations between the ELMo and BERT embeddings and the more classical static word embeddings by adding as targets \textsc{Dist\_ngh} the most similar neighbors to the keys of our study retrieved with the \emph{cosine} measure by a Skip-gram model trained on a large corpus.

This investigation of the semantic properties of ELMo and BERT can be viewed as a kind of semantic probing task and related to the work of \citet{schick-aaai-20} and their \emph{WordNet Language Model Probing}. While their overall objective is different from ours, their probing task is also significantly different since their notion of pattern is more adapted to syntagmatic relations than to paradigmatic relations.

\begin{table*}[t]
\centering
{\small
\begin{tabular}{lccccccccccccc}
\toprule
 &  &  & &\multicolumn{3}{c}{\textbf{ELMo}} & &\multicolumn{6}{c}{\textbf{BERT}} \\ 
 \cmidrule(lr){5-7} \cmidrule(lr){9-14}   
 & \textbf{avg. \#tgt} & \textbf{random} & &\textbf{L0} & \textbf{L1} & \textbf{L2} & &\textbf{L1} & \textbf{L3} & \textbf{L5} & \textbf{L8} & \textbf{L10} & \textbf{L12}\\   \midrule
\textsc{Syn} & 2.1 & 5.3 & &30.9 & 29.3 & 26.9 & &33.5  & 34.5  & 36.1  & \textbf{36.7}  & 36.0 & 35.1 \\ 
\textsc{Hype} & 1.9 & 6.7 & &4.3 & 6.1 & 6.2 & &7.8  & 9.3  & 9.9  & 10.9  & 11.1 & \textbf{11.2} \\ 
\textsc{Hypo} & 5.9 & 14.5 & &11.4 & 13.8 & \textbf{14.8} & &10.5  & 11.1  & 11.7  & 12.3  & 11.9 & 12.3 \\ 
\textsc{Cohyp} & 8.3 & 29.5 & &6.7 & 7.9 & \textbf{9.6} & &6.4  & 6.7  & 7.5  & 7.7  & 7.8 & 7.9 \\ 
\textsc{Dist\_ngh} & 10 & 44.0 & &\textbf{46.6} & 42.9 & 42.5 & &41.7  & 38.4  & 34.7  & 32.3  & 33.1 & 33.5 \\ \bottomrule
\end{tabular}
}
\caption{P@1 ($\times100$) for the ranking, based on SemCor's sentences, of the reference targets for a set of keys\footnotemark.}
\label{tab:semcor}
\end{table*}

\section{Experiments: Study of Contextual Word Embeddings}\label{sec:exp_semcor}

\subsection{Experimental Setup}

The implementation of the principles of the previous section is associated with some choices about  test sentences and semantic relations between keys and targets. First, both keys and targets are restricted to nouns. The number of targets for each key is limited to 40 to have comparable results for all keys. Moreover, we only retain keys having targets for all the five types of relations we consider, once again for the homogeneity of results. We also limit the number of targets for each type of relation to 10, with an additional limit of 30 for all WordNet relations. In practice, this limit only concerns hyponymy and cohyponymy relations, whose number tends to be high. The first column of Table~\ref{tab:semcor} gives the average number of targets of each key according to their type.

For WordNet relations, we adopt the following definitions for each type of target:

\begin{compactitem}
 \item synonyms [\textsc{Syn}]: all the words of $Synset_{key}$, the synset of the sense associated with the occurrence of the key in a test sentence;
 \item hypernyms [\textsc{Hype}]: all the words of the synsets $Synset_{hype}$ having a direct hypernymy relation with $Synset_{key}$;
 \item hyponyms [\textsc{Hypo}]: all the words of the synsets having a direct hyponymy relation with $Synset_{key}$;
 \item cohyponyms [\textsc{Cohyp}]: all the words of the synsets, except $Synset_{key}$, having a direct hyponymy relation with the $Synset_{hype}$ synsets.
\end{compactitem}

As mentioned previously, the \textsc{Dist\_ngh} targets are obtained by a Skip-gram model, trained on a 1 billion word subset of the Annotated English Gigaword corpus \cite{napoles-naacl_ws-12} with the best hyperparameter values from \cite{baroni-acl-14}. More precisely, we use this model to retrieve the 10 first neighbors, among a vocabulary of \numprint{20813} nouns, of each key that are not present in the set of WordNet targets. In that case, this selection is not done according to the sense of the key in a test sentence since we do not have access to word senses.

Concerning test sentences, an upper limit on the number of sentences for each key is also fixed: no more than 20 sentences for each sense of a key. Finally, our evaluation is based on \numprint{41079} SemCor's sentences, which represents around \numprint{4.5} sentences by sense on average and \numprint{7.9} sentences for each of the \numprint{5241} keys. These keys cover a large spectrum of frequencies since if we refer to the subpart we use of the Gigaword corpus, the most frequent key, \emph{year}, has \numprint{2991899} occurrences while the least frequent keys such as \emph{inadvertence} have only 22 occurrences.

\subsection{Evaluation}

\footnotetext{The percentage of cases in which the first target of a key corresponds to a paradigmatic relation is equal to $100 - P@1(\textsc{Dist\_ngh})\times100$. For instance, 56\% for the random ranker.} The results of the ranking of the targets selected for all our keys are presented in Table~\ref{tab:semcor} for each layer of ELMo and BERT and each type of target. Moreover, the second column provides the P@1 values of a random ranker (average values over 100 runs). P@1 in this context corresponds to the proportion of sentences that rank first a target linked to a key with a specific type of relation (one by row). 

This table first shows that the different layers of ELMo and BERT do not have exactly the same semantic properties, which is a confirmation of previous findings as those of \citet{ethayarajh-emnlp-19} or \citet{wu-acl-20}. It also shows some similarities and differences between ELMo and BERT. The most obvious difference is that BERT has much higher results than ELMo for synonyms and hypernyms but lower results for hyponyms and cohyponyms, meaning that BERT favors semantic similarity over semantic relatedness \citep{budanitsky-cl-06} as we can consider that hyponyms and cohyponyms, despite their paradigmatic nature, are less strong than synonyms and hypernyms in terms of semantic similarity. The trend is even more obvious for the \textsc{Dist\_ngh} relations coming from the static embeddings, which are more likely to be syntagmatic relations.

ELMo and BERT also exhibit some differences in their layers. While the capacity of ELMo to rank synonyms first steadily decreases as we consider higher layers, it first increases until layer~8 for BERT, then tends to decrease. This observation has also to be put in relation to the results of \citet{ethayarajh-emnlp-19}, who observed that word representations are more context-specific as the level of their layers increases. From the viewpoint of ELMo, it means that having more contextual representations is likely to favor semantic relatedness over semantic similarity. This is not unexpected since from the semantic viewpoint, contextual relations are syntagmatic rather than paradigmatic relations. This result is less obvious for BERT since its first layers have the best precision for \textsc{Dist\_ngh} relations but a change more compatible with the results of ELMo occurs after layer~8 concerning the ranking of synonyms and \textsc{Dist\_ngh} relations.

However, ELMo and BERT also have strong convergences. First, the most significant effect of the ranking by ELMo and BERT is obtained for synonyms. P@1 is up to nearly six times higher for the best ELMO's layer and nearly seven times for the best BERT's layer compared to the random ranking, which indicates that the application of the distributional hypothesis as described in Section~\ref{sec:distr_hyp} is an interesting method for identifying the synonyms of a word among a list of its distributional neighbors. It also illustrates the fact that even if some differences can be noted between  layers in terms of semantic orientation, they all have a strong bias towards synonymy. We can also observe that both ELMo and BERT have a strong inverse correlation between P@1 for synonyms and P@1 for \textsc{Dist\_ngh} relations: when a layer tends to favor strict semantic similarity, it logically obtains worse results for semantic relatedness at the same time. However, this correlation does not lead to results for \textsc{Dist\_ngh} relations much lower than the results of a random ranker for ELMo, which suggests that ELMo is not radically different from static embeddings from the viewpoint of the semantic similarity it conveys. This trend is less clear-cut for BERT since its results for synonyms are more comparable to those for \textsc{Dist\_ngh} relations but BERT nevertheless gives significant importance to \textsc{Dist\_ngh} relations. A closer look at the first ranked \textsc{Dist\_ngh} word also shows that it frequently has a strong semantic relationship with the key. In some cases, it is one of its synonyms but for a different sense, that is close to the sense of the current occurrence. For instance, in the sentence:

\begin{adjustwidth}{.3cm}{0cm}
    \setlength{\parskip}{6pt}
    \small
Land reform programs need to be supplemented with programs for promoting rural credits and [...] in \textbf{agriculture}.
\end{adjustwidth}

{\setlength{\parskip}{6pt}
\noindent the word ranked first by the first layer of ELMo, \emph{farming}, is a synonym of the second sense of \emph{agriculture} -- \emph{the practice of cultivating the land or raising stock} -- whereas this occurrence is tagged in SemCor with its first sense -- \emph{a large-scale farming enterprise}. In some other cases, the top word is actually a synonym of the key but is not considered as such in WordNet. For instance, the top word \emph{capability} for the key \emph{ability}. Finally, it is also frequent to find as the top word an antonym of the key, as \emph{inaction} for the key \emph{action}. While this is not an intended outcome, antonyms are known to be distributionally very similar to synonyms and their presence confirms the observed trend to favor semantic similarity.}

\begin{table*}[t]
\centering
{\small
\begin{tabular}{lccccccccccccc}
\toprule
 &  \textbf{avg.} &  & &\multicolumn{3}{c}{\textbf{ELMo}} & &\multicolumn{6}{c}{\textbf{BERT}} \\ 
 \cmidrule(lr){5-7} \cmidrule(lr){9-14}   
 & \textbf{\#ref. rel.} & \textbf{Skip-gram} & &\textbf{L0} & \textbf{L1} & \textbf{L2} & &\textbf{L1} & \textbf{L3} & \textbf{L5} & \textbf{L8} & \textbf{L10} & \textbf{L12}\\   \midrule
\textsc{Syn} & 3.5 & 18.6 &  & 22.6 & \textbf{23.0} & 21.4 &  & 20.1 & 20.6 & 20.9 & 21.4 & 21.3 & 21.9\tabularnewline
\textsc{Hype} & 5.0 & 6.9 &  & 5.8 & 6.6 & 6.3 &  & 8.1 & 8.3 & 8.6 & 8.9 & \textbf{9.0} & 8.5\tabularnewline
\textsc{Hypo} & 10.8 & 11.8 &  & 13.3 & \textbf{14.2} & 13.7 &  & 10.5 & 10.8 & 11.3 & 11.5 & 10.9 & 10.2\tabularnewline
\textsc{Cohyp} & 56.3 & 27.9 &  & 33.5 & \textbf{34.8} & 33.0 &  & 30.0 & 30.7 & 31.2 & 31.3 & 31.2 & 31.6\tabularnewline 
 \bottomrule
\end{tabular}
}
\caption{P@1 ($\times100$) for the ranking of Skip-gram's distributional neighbors by contextual models.}
\label{tab:rerank_layer}
\end{table*}

\section{Contextual Word Embeddings vs Static Embeddings}

\subsection{Principles}\label{sec:rerank_princ}

Besides the investigation of the semantic properties of contextual word embeddings, the method we have presented can also be adapted to study the differences between contextual and static word embeddings concerning these properties. More precisely, the idea is to define targets by replacing WordNet semantic relations with relations characterizing static embeddings. Due to the distributional nature of these embeddings, we choose to associate as targets with each key, corresponding to a word $W_i$, its most similar distributional neighbors according to the considered static embeddings. As previously, targets are reranked according to a set of test sentences and we use paradigmatic relations in WordNet for determining a posteriori which types of relations contextual word embeddings favor compared to static embeddings. 

However, since static embeddings are defined at the word type level, their comparison with contextual word embeddings requires aggregating the contextual representations of keys and targets built from test sentences for producing type level representations. We classically distinguish between early and late fusion. The early fusion consists in this context in aggregating the contextual representations of a key or a target extracted from the encoding by ELMo or BERT of the test sentences where it occurs. For this aggregation, we consider three operators \cite{bommasani-acl-20}: \emph{average}, \emph{element-wise maximum}, and \emph{minimum}. The late fusion approach \citep{curran-emnlp-02} produces a reranking of the neighbors of a word for each test sentence and merges the resulting rankings according to methods that are typically used in Information Retrieval for merging ranked lists of retrieved documents. More precisely, we experiment with two kinds of methods: the \emph{Borda}, \emph{Condorcet} \citep{nuray-ipm-06} and \emph{Reciprocal Rank} (RRF) \citep{cormack-sigir-09} fusions based on ranks and the \emph{CombSum} fusion based on similarity values, normalized with the Zero-one method \citep{wu-air-06}.

The definition of static embeddings at the word level also influence the way test sentences are selected: a static embedding is not associated with a particular sense of a word but, according to the idea developed by \citet{mccarthy-acl-04} that most words have one predominant sense in a specific corpus, it is supposed to be related to the predominant sense of the word it is associated with in the corpus used for its building. Hence, we adopt a strategy for selecting test sentences accordingly. Its first step consists in selecting randomly for each word $W_i$ a large enough set of sentences containing the word, at most $N_{sent}$, from the corpus used for retrieving the targets. The resulting set of contexts for each word can be considered  statistically representative of its various senses.  

The second step aims at selecting a smaller number of context sentences for performing the reranking at a reasonable cost while taking into account the predominant sense hypothesis. Following this hypothesis, we assume that most of the test sentences first selected for a word refer to its predominant sense. As a consequence, averaging the representations of the word resulting from the encoding of the set $\mathcal{C}$ = $\{S_j\}$ of its test sentences by ELMo or BERT should lead to a representation of the word, denoted $v_{(W_i, \mathcal{C})}$, very close to its predominant sense. Considering that we select a fixed number $N_{\mathcal{C}}$ of test sentences, we propose the following options:

\begin{compactitem}
\item \emph{random}: it is our base option in which $N_{\mathcal{C}}$ sentences are randomly selected among the $N_{sent}$ initially selected for the word;
 \item \emph{closest\_avg}: we select test sentences $S_j$ such that the representation of the word $v_{(W_i, S_j)}$ is closest to $v_{(W_i, \mathcal{C})}$, with the idea to favor the homogeneity among the test sentences towards the predominant sense of the word; 
 \item \emph{farthest\_avg}: this is the opposite of \emph{closest\_avg}. We select test sentences such that $v_{(W_i, S_j)}$ is farthest to $v_{(W_i, \mathcal{C})}$ to increase the presence of minor senses of the word;
 \item \emph{uniform}: the idea is to account for the diversity of word's senses by selecting $N_{\mathcal{C}}$ that are uniformly distributed in terms of the similarity of $v_{(W_i, S_j)}$ to $v_{(W_i, \mathcal{C})}$. 
\end{compactitem}

\subsection{Experimental Setup}

The main difference with the setup of our first experiment concerns test sentences: they are  selected from the corpus used for training the static embeddings and keys are not semantically disambiguated in them. More precisely, we select 10 test sentences for each key with a size between 10 and 90 words for having a significant and focused context. For the static embeddings, we rely on the same Skip-gram model as the one used for extracting the \textsc{Dist\_ngh} relations. In a first experiment (see Section~\ref{sec:static_eval1}), we apply the reranking process to the same \numprint{5241} keys as in Section~\ref{sec:exp_semcor} for having a direct comparison with Table~\ref{tab:semcor}. Test sentences are selected randomly and we apply an early fusion by averaging the representations extracted from test sentences. In a second experiment (see Section~\ref{sec:static_eval2}), we consider a larger number of \numprint{10302} keys for testing the various options presented in the previous section. In both cases, the targets correspond to the first 10 distributional neighbors of the keys, retrieved by the \emph{cosine} measure. 

As in \citep{piasecki-gwc-18}, our gold standard neighbors are obtained by extracting from WordNet the words linked to a key through the same types of relations as in Section~\ref{sec:exp_semcor} but the number of relations is larger since we consider all the synsets in which the key is present (see the first column of Table~\ref{tab:rerank_layer}). We report the precision at the first rank (P@1) of the retrieved neighbors for each type of relation in the first experiment and add precisions at ranks 2 and 5 (P@2 and P@5) for the second experiment but without detailing these measures according to the different types of relations.

\subsection{Skip-gram versus Contextual Models}\label{sec:static_eval1}

Table~\ref{tab:rerank_layer} compares the distributional neighbors retrieved by the Skip-gram model with their ranking by ELMo and BERT according to the methodology we propose. The first observation is that while these models significantly\footnote{Differences are judged significant according to a paired Wilcoxon test if $p \leq 0.01$.} favor synonyms and cohyponyms compared to Skip-gram, the situation is more complex for hypernyms and hyponyms: ELMo degrades Skip-gram's results for hypernyms but improves them for hyponyms while BERT does exactly the opposite. We can also note that the Skip-gram model largely favors cohyponymy over the other lexical relations, a trend that tends to be increased by ELMo and BERT, to a greater extent by ELMo. Interestingly, in Table~\ref{tab:semcor}, the ranking of cohyponyms by ELMo and BERT is much worse than the one obtained by a random ranker. This difference illustrates the contextual nature of these models but also its limits. 
In the experiment of Section~\ref{sec:exp_semcor}, keys are semantically disambiguated in test sentences and targets are related to their sense. In such a situation, a contextual model can favor the targets most semantically linked to their key, such as synonyms, because they produce representations that are specific to the considered context, which is related to the sense of the key. On the contrary, in this second experiment, the targets cover all the senses of the key and in such a configuration, corresponding to the word type level, the representations built by contextual models favor semantic relatedness over semantic similarity and are closer to static embeddings, even if they globally tend to outperform them. This effect has a much greater impact for synonyms in the case of BERT than for ELMo, which probably results from a greater sensitivity of BERT to the context than ELMo. However, it does not modify the pattern of results among BERT's layers, with best results around layer 8, while ELMo's best results are fairly surprisingly obtained by a more contextualized layer than previously. We can note that the observation about BERT is close to the findings of \citet{chronis-conll-20}, who found that BERT's layer 7 ``is optimal for estimating similarity'' in their comparison with relatedness.

Table~\ref{tab:rerank_layer} also shows that BERT is globally closer to the Skip-gram model than ELMo from the viewpoint of paradigmatic relations, except for hypernyms. This is surprising if we consider the results of Table~\ref{tab:semcor}, where BERT globally outperforms ELMo for paradigmatic relations, especially for synonyms. We analyze this observation as a consequence of the use of wordpieces by BERT for dealing with out-of-vocabulary words. In the experiments of Section~\ref{sec:exp_semcor}, 73\% of words are part of BERT's vocabulary while this ratio decreases to 49.4\% in the experiments of Table~\ref{tab:rerank_layer}. Following \citet{bommasani-acl-20} and others, we build the representation of a word split into wordpieces by averaging the representations of its wordpieces. While this is considered as the best strategy in such a situation, our results show that it has a significant negative impact when the number of words split into wordpieces reaches a certain level and introduces a form of bias into results as all words do not have the same status.

\begin{table}[t]
\centering
{\small
\begin{tabular}{lcccc}
\toprule 
 & \textbf{P@1}{ } & \textbf{P@2}{ } & \textbf{P@5}{ }\tabularnewline
\midrule 
{n=10, s=10 } & \textbf{41.8}{ } & {34.5 } & {24.0 }\tabularnewline
\midrule 
{s=10, n=5 } & {41.3$^{\dag}$ } & {33.6 } & {21.6 }\tabularnewline
{s=10, n=15 } & {41.6$^{\dag}$ } & \textbf{34.8}{$^{\dag}$ } & {24.7 }\tabularnewline
{n=15, s=5 } & {41.2$^{\dag}$ } & {34.6$^{\dag}$ } & \textbf{24.8}{ }\tabularnewline
{n=15, s=15 } & {41.6$^{\dag}$ } & \textbf{34.8}{$^{\dag}$ } & \textbf{24.8}{ }\tabularnewline
\bottomrule
\end{tabular}
}
\caption{Impact of the number of neighbors ($n$) and test sentences ($s$) for ELMo$_{L1}$ (reference configuration of Table~\ref{tab:rerank_layer}: n=10, s=10; $^{\dag}$: non-significant difference compared to this reference).}
\label{tab:rerank_sent_ngh}
\end{table}

\subsection{Impact of Reranking Options}\label{sec:static_eval2}

We have seen in Section~\ref{sec:rerank_princ} that the method we propose depends on the definition of several hyperparameters and options. Table~\ref{tab:rerank_sent_ngh} shows the impact of two parameters of the method that influence both its performance and speed: the number of test sentences ($s$) and the number of reranked neighbors ($n$). 
The evaluation of this impact is done according to ELMo's layer~1, which globally obtains the best results in reranking targets from Skip-gram, as indicated by Table~\ref{tab:rerank_layer}. Moreover, it is also performed by merging the four types of semantic relations we consider in Table~\ref{tab:rerank_layer}. 
We can globally observe that the proposed method is not very sensitive to changes in the value of these two parameters as their impact on results is generally not significant. These experiments also show that reranking only a small number of neighbors under-exploits the capabilities of the method, especially when the number of test sentences is small. However, it is interesting to note that the proposed method can be effective with only a restricted number of test sentences. Finally, we adopt s=10 and n=15 as the more interesting compromise between the quality of results and the speed of the method for our following evaluations.

In Section~\ref{sec:rerank_princ}, we have seen that considering keys and targets at word level requires aggregating token level representations, with several possible options. Taking as reference the \emph{average} option adopted in the evaluation of Section~\ref{sec:static_eval1}, Table~\ref{tab:rerank_fus} shows that while the best results are obtained by the \emph{average} option, an early fusion approach, the late fusion approaches globally outperform the other early fusion options. However, the results of the different settings are fairly homogeneous. This is particularly true for all the results of the late fusion approaches, which suggests that for a key or a target, the rerankings produced by the test sentences are very similar.  

The last evaluation of the options of the reranking concerns the way test sentences are selected from their initial set. As for the aggregation of token representations, the differences between the tested options reported in Table~\ref{tab:rerank_sent_sel} are not very high. Only the \emph{farthest\_avg} option is significantly worse than \emph{random}, our reference, which is not very  surprising since favoring the most atypical test sentences does not seem a priori a good option. Finally, the \emph{uniform} method appears as the best choice both because it obtains the best results, even if they are not statistically different from those of \emph{random}, and more importantly, it is deterministic, which is not the case of \emph{random}.

\begin{table}[t]
\centering
{\small
\begin{tabular}{lcccc}
\toprule 
 & \textbf{P@1}{ } & \textbf{P@2}{ } & \textbf{P@5}{ }\tabularnewline
\midrule 
{average } & \textbf{41.6}{ } & \textbf{34.8}{ } & \textbf{24.7}{ }\tabularnewline
{max } & {40.6 } & {33.9 } & {24.3 }\tabularnewline
{min } & {40.5 } & {33.9 } & {24.3 }\tabularnewline
\midrule 
{Borda } & {41.1 } & {34.4 } & {24.6 }\tabularnewline
{Condorcet } & {40.9 } & {34.4 } & {24.6 }\tabularnewline
{RRF } & {41.1$^{\dag}$ } & {34.4 } & {24.6 }\tabularnewline
{CombSum } & {41.0 } & {34.3 } & {24.6$^{\dag}$ }\tabularnewline
\bottomrule
\end{tabular}
}
\caption{Impact of the fusion method for token level representations with ELMo$_{L1}$ (reference method in Table~\ref{tab:rerank_sent_ngh}: \emph{average}). }
\label{tab:rerank_fus}
\end{table}

\begin{table}[t]
\centering
{\small
\begin{tabular}{lcccc}
\toprule 
 & \textbf{P@1}{ } & \textbf{P@2}{ } & \textbf{P@5}{ }\tabularnewline
\midrule 
{random } & {41.6 } & {34.8 } & {24.7 }\tabularnewline
{closest\_avg } & {41.4$^{\dag}$ } & {34.5 } & \textbf{24.8}{$^{\dag}$ }\tabularnewline
{farthest\_avg } & {39.9 } & {33.6 } & {24.1 }\tabularnewline
{uniform } & \textbf{41.8}{$^{\dag}$ } & \textbf{34.9}{$^{\dag}$ } & \textbf{24.8}{$^{\dag}$ }\tabularnewline
\bottomrule
\end{tabular}
}
\caption{Impact of the selection method of test sentences for ELMo$_{L1}$ (reference method in Table~\ref{tab:rerank_sent_ngh}: \emph{random}).}
\label{tab:rerank_sent_sel}
\end{table}

\begin{table}[t]
\centering
{\small
\begin{tabular}{lcccc}
\toprule 
 & \textbf{P@1}{ } & \textbf{P@2}{ } & \textbf{P@5}{ }\tabularnewline
\midrule 
{initial$_{high}$ } & {41.6 } & {35.3 } & {26.4 }\tabularnewline
{initial$_{low}$ } & {30.0 } & {24.1 } & {16.8 }\tabularnewline
\midrule 
{reranking$_{high}$ } & {50.8 } & {43.1 } & {31.1 }\tabularnewline
{reranking$_{low}$ } & {32.8 } & {26.6 } & {18.4 }\tabularnewline
\bottomrule
\end{tabular}
}
\caption{Comparison of the \emph{initial} ranking of the targets given by Skip-gram and their \emph{reranking} by ELMo$_{L1}$ (corresponding to \emph{uniform} in Table~\ref{tab:rerank_sent_sel}) according to the frequency of keys (\emph{high} or \emph{low}).}
\label{tab:rerank_high_low}
\end{table}

\begin{table*}[th]
\centering
\begin{adjustbox}{max width=\textwidth}
{\small
\begin{tabular}{l>{\raggedright}m{0.5\textwidth}>{\raggedright}m{0.5\textwidth}}
\toprule 
\textbf{Keys} & \textbf{Initial order of targets} & \textbf{Targets after reranking}\tabularnewline
\midrule
response & reaction$_{syn}$, wake$_{coh}$, action, criticism, rebuke, condemnation, denunciation$_{coh}$, explanation$_{coh}$, comment$_{coh}$ & reaction$_{syn}$, reply$_{syn}$, answer$_{syn}$, counteraction, rebuke, explanation$_{coh}$, unresponsiveness, condemnation, action\tabularnewline
\midrule 
bureau & department, telephone, reporting, newsroom, enumerator, agency$_{syn}$, office$_{syn}$,  correspondent, authority & agency$_{syn}$, department, office$_{syn}$, authority$_{syn}$, newsroom, correspondent, enumerator, chief, deputy\tabularnewline
\midrule 
supposition & contrary, conjecture$_{syn}$, theory$_{hype}$, fact, notion, assumption$_{syn}$, characterisation, assertion, reductionism & assumption$_{syn}$, assertion, conjecture$_{syn}$, hypothesis$_{syn}$, notion, belief, surmise$_{syn}$, theory$_{hype}$, characterisation\tabularnewline
\midrule
lookout & alert, loiterer, vigilance, binoculars, spotter$_{syn}$, prowl, watch$_{syn}$, warning, rubbernecker, sentry$_{syn}$ & prowl, sentry$_{syn}$, spotter$_{syn}$, sentinel$_{syn}$, watch$_{syn}$, sightseer, alert, picnicker, vigilance\tabularnewline
\bottomrule 
\end{tabular}
}
\end{adjustbox} 
\caption{Examples of reranking by ELMo$_{L1}$ of the first targets of some keys (\emph{syn}: synonym, \emph{coh}: cohyponym, \emph{hype}: hypernym).}
\label{tab:examp}
\end{table*}

\subsection{Complementary Analyses and Examples}

While Table~\ref{tab:rerank_layer} shows the differences between static and contextual embeddings according to the type of semantic relation, it is also interesting to perform such a comparison according to the frequency of keys. Table~\ref{tab:rerank_high_low} illustrates it for ELMo by the split of results into two equally balanced frequency slices -- \emph{high} and \emph{low}. While the reranking method leads to an improvement for the two slices, this improvement is clearly more significant for high-frequency words. Since we use the same number of test sentences for all words, we assume that this finding is not linked to the number of occurrences of the words but their number of senses. As a contextual model, ELMo produces more specific representations than the Skip-gram model we use for retrieving the targets and these focused representations are more effective for retrieving relevant neighbors in terms of paradigmatic relations, especially for polysemous words such as high-frequency words.

Finally, Table~\ref{tab:examp} presents more qualitatively for some keys the differences between the first targets of the Skip-gram model and their reranking by ELMo. While it shows the global tendency of ELMo to favor synonymy more than Skip-gram does, it also illustrates, in accordance with the results of Table~\ref{tab:semcor}, that this feature is particularly effective when a significant proportion of the reranked targets are linked to their key with paradigmatic relations.

\section{Conclusion and Perspectives}

In this article, we have investigated how the distributional principle of substitution can be used as a form of probe for testing the kind of semantic similarity conveyed by ELMo and BERT. Experiments with WordNet paradigmatic relations and SemCor at the level of word senses have shown that the contextual nature of these models clearly favors synonymy in terms of lexical relations. Moreover, we have adapted the same method for studying the differences between contextual and static embeddings, with the conclusion that the bias of contextual embeddings towards semantic similarity at the token level is reduced at the word type level but is still present. 

One extension of this work is to address the problem raised by the representation of words split into wordpieces by BERT. More precisely, we plan to test  two types of solutions: one is a specific mechanism for building word representations from their wordpieces as the One-Token Approximation of \citet{schick-aaai-20}; the other relies on character-based models such as the recent CharacterBERT \citep{el-boukkouri-coling-20} or CharBERT \citep{ma-coling-20} models.

\section*{Acknowledgments}

This work has been partially funded by French National Research Agency (ANR) under project ADDICTE (ANR-17-CE23-0001).

\bibliography{biblio}
\bibliographystyle{acl_natbib}

\appendix
\newpage

\section*{Detailed Parameters, Tools and Resources}

Parameters of \texttt{word2vec}\footnote{\url{https://code.google.com/archive/p/word2vec}} for the Skip-gram model used for building our static embeddings:

\begin{compactitem}
\item minimal count = 5
\item vector size = 300
\item window size = 5
\item number of negative examples = 10
 \item subsampling probability of the most frequent words = $10^{-5}$
\end{compactitem}

For ELMo, our experiments make use of the Original (5.5GB) pre-trained ELMo model with 93.6 million parameters available at: \url{https://allennlp.org/elmo}. For BERT, our experiments relied on the \texttt{bert-base-uncased} model with 110 million parameters available at: \url{https://storage.googleapis.com/bert_models/2018_10_18/uncased_L-12_H-768_A-12.zip}. All our experiments were performed on a node of a cluster with 4 GTX 1080 GPU cards and 10GB RAM each.

\end{document}